\documentclass[conference]{IEEEtran}
\IEEEoverridecommandlockouts
\usepackage{cite}
\usepackage{amsmath,amssymb,amsfonts}
\usepackage{algorithmic}
\usepackage{graphicx}
\usepackage{textcomp}
\usepackage{xcolor}
\def\BibTeX{{\rm B\kern-.05em{\sc i\kern-.025em b}\kern-.08em
    T\kern-.1667em\lower.7ex\hbox{E}\kern-.125emX}}

\usepackage{booktabs}

\newcommand\blfootnote[1]{%
  \begingroup
  \renewcommand\thefootnote{}\footnote{#1}%
  \addtocounter{footnote}{-1}%
  \endgroup
}

\usepackage{amstext} 
\usepackage{array}   
\newcolumntype{L}{>{$}l<{$}} 
\newcolumntype{R}{>{$}r<{$}} 

\usepackage{caption}
\usepackage{subcaption}
\usepackage{makecell}
\usepackage{multirow}
\usepackage{mathtools}

\usepackage{hyperref}

\usepackage{amsmath,amssymb,amsthm} 
\usepackage[algoruled,vlined,nofillcomment,linesnumbered]{algorithm2e}

\usepackage[textsize=footnotesize,color=green!40]{todonotes}

\newcommand\numberthis{\addtocounter{equation}{1}\tag{\theequation}}

\newcommand{\cX}{\mathcal{X}}
\newcommand{\cY}{\mathcal{Y}}
\newcommand{\cW}{\mathcal{W}}
\newcommand{\R}{\mathbb{R}}
\renewcommand{\Pr}{\mathsf{Pr}}
\newcommand{\norm}[1]{\| #1 \|}
\newcommand{\argmin}{\mathop{argmin}}

\newcommand{\defacto}{\emph{de facto}}

\newcommand{\eg}{\emph{e.g.}}

\newcommand{\ie}{\emph{i.e.}}
\newcommand{\vs}{\emph{vs.}}

\usepackage{acronym}
\acrodef{DP}{Differential Privacy}
\acrodef{LDP}{Local \ac{DP}}
\acrodef{mDP}{Metric \ac{DP}} 
\acrodef{NLP}{Natural Language Processing}
\acrodef{ML}{Machine Learning}
\acrodef{AI}{Artificial Intelligence}
\acrodef{biLSTM}{Bidirectional LSTM} 
\acrodef{GESD}{Geometric mean of Euclidean and Sigmoid Dot product}
\acrodef{MAP}{Mean Average Precision}
\acrodef{MRR}{Mean Reciprocal Rank}
\acrodef{PII}{\emph{personally identifiable information}}
\acrodef{MH}{Metropolis-Hastings}

\newcommand{\LP}{\left(}
\newcommand{\LB}{\left[}
\newcommand{\RP}{\right)}
\newcommand{\RB}{\right]}
\newcommand{\LN}{\left\|}
\newcommand{\RN}{\right\|}
\newcommand{\LA}{\left\langle}
\newcommand{\RA}{\right\rangle}

\newcommand{\Lcal}{\mathcal{L}}

\newcommand{\Reals}{\mathbb{R}}
\newcommand{\Spherical}{\mathbb{S}}
\newcommand{\Hyperbolic}{\mathbb{H}}

\newcommand{\arcosh}{\operatorname{arcosh}}
\newcommand{\hg}{{}_2F_1}

\newtheorem{lemma}{Lemma}
\newcommand{\pc}{Poincar{\'e}}

\begin{document}

\title{Leveraging Hierarchical Representations\\
for Preserving Privacy and Utility in Text
}


\author{\IEEEauthorblockN{Oluwaseyi Feyisetan}
\IEEEauthorblockA{\textit{Amazon} \\
\texttt{sey@amazon.com}}
\and
\IEEEauthorblockN{Tom Diethe}
\IEEEauthorblockA{\textit{Amazon} \\
 \texttt{tdiethe@amazon.co.uk}}
\and
\IEEEauthorblockN{Thomas Drake}
\IEEEauthorblockA{\textit{Amazon} \\
 \texttt{draket@amazon.com}}
}

\maketitle

\begin{abstract}
Guaranteeing a certain level of user privacy in an arbitrary piece of text is a challenging issue. However, with this challenge comes the potential of unlocking access to vast data stores for training machine learning models and supporting data driven decisions. We address this problem through the lens of $d_{\chi}$-privacy, a generalization of Differential Privacy to non Hamming distance metrics. In this work, we explore word representations in Hyperbolic space as a means of preserving privacy in text. We provide a proof satisfying $d_{\chi}$-privacy, then we define a probability distribution in Hyperbolic space and describe a way to sample from it in high dimensions. Privacy is provided by perturbing vector representations of words in high dimensional Hyperbolic space to obtain a semantic generalization. We conduct a series of experiments to demonstrate the tradeoff between privacy and utility. Our privacy experiments illustrate protections against an authorship attribution algorithm while our utility experiments highlight the minimal impact of our perturbations on several downstream machine learning models. Compared to the Euclidean baseline, we observe $> 20$x greater guarantees on expected privacy against comparable worst case statistics.
\end{abstract}

\begin{IEEEkeywords}
privacy; document redaction; data sanitization
\end{IEEEkeywords}

\section{Introduction}
\blfootnote{Accepted at ICDM 2019} In \ac{ML} tasks and \ac{AI} systems, training data often consists of information collected from users. This data can be sensitive; for example, in conversational systems, a user can explicitly or implicitly disclose their identity or some personal preference during their voice interactions. Explicit \ac{PII} (such as an individual's PIN or SSN) can potentially be filtered out via rules or pattern matching. However, more subtle privacy attacks occur when seemingly innocuous information is used to discern the private details of an individual \cite{dwork2017exposed}. This can lead to a number of attack vectors -- ranging from human annotators making deductions on user queries \cite{feyisetan2019privacy} to membership inference attacks being launched against machine learning models that were trained on such data \cite{shokri2017membership}. As a result, privacy-preserving analysis has increasingly been studied in statistics, machine learning and data mining \cite{dwork2006calibrating,shokri2015privacy} to build systems that provide better privacy guarantees.

Of particular interest are these implicit, subtle privacy breaches which occur as a result of an adversary's ability to leverage observable patterns in the user's data. These \emph{tracing attacks} have been described to be akin to `fingerprinting' \cite{bun2018fingerprinting} due to their ability to identify the presence of a user's data in the absence of explicit \ac{PII} values. 
The work by \cite{songshmatikovkdd} demonstrates how to carry out such tracing attacks on \ac{ML} models by determining if a user's data was used to train the model. These all go to illustrate that the traditional notion of \ac{PII} which is used to build anonymization systems is fundamentally flawed \cite{schwartz2011pii}. Essentially, any part of a user's information can be used to launch these attacks, and we are therefore in a post-\ac{PII} era \cite{dwork2017exposed}. This effect is more pronounced in naturally generated text as opposed to statistical data where techniques such as \ac{DP} have been established as a \defacto{} way to mitigate these attacks.

While providing quantifiable privacy guarantees over a user's textual data has attracted recent attention \cite{Coavoux2018PrivacypreservingNR,Weggenmann_2018}, there is significantly more research into privacy-preserving statistical analysis. In addition, most of the text-based approaches have focused on providing protections over vectors, hashes and counts \cite{erlingsson2014rappor,wang2017locally}. The question remains: what quantifiable guarantees can we provide over the actual text? We seek to answer that question by adopting the notion of $d_{\chi}$-privacy \cite{andres2013geo,chatzikokolakis2013broadening,chatzikokolakis2015constructing}, an adaptation of \ac{LDP} \cite{kasiviswanathan2011can} which was designed for providing privacy guarantees over location data. $d_{\chi}$-privacy generalizes \ac{DP} beyond Hamming distances to include Euclidean, Manhattan and Chebyshev metrics, among others. In this work, we demonstrate the utility of the Hyperbolic distance for $d_{\chi}$-privacy in the context of textual data. This is motivated by the ability to better encode hierarchical and semantic information in Hyperbolic space than in Euclidean space \cite{krioukov2010hyperbolic,nickel2017poincare,nickel2018learning}.

At a high level, our algorithm preserves privacy by providing \emph{plausible deniability} \cite{bindschaedler2017plausible} over the contents of a user's query. We achieve this by transforming selected words to a high dimensional vector representation in Hyperbolic space as defined by \pc{} word embeddings \cite{nickel2017poincare}. We then perturb the vector by sampling noise from the same Hyperbolic space with the amount of added noise being proportional to the privacy guarantee. This is followed by a post-processing step of \emph{discretization} where the noisy vector is mapped to the closest word in the embedding vocabulary. This algorithm conforms to the $d_{\chi}$-privacy model introduced by \cite{andres2013geo} with our transformations carried out in higher dimensions, in a different metric space, and within a different domain. To understand why this technique preserves privacy, we describe motivating examples in Sec.~\ref{sec:mech_overview} and define how we quantify privacy loss by using a series of interpretable proxy statistics in Sec.~\ref{sec:privacy_calibration}.

\subsection{Contributions}
Our contributions in this paper are summarized as follows:
\begin{enumerate}
\item We demonstrate that the Hyperbolic distance metric satisfies $d_{\chi}$-privacy by providing a formal proof in the Lorentz model of Hyperbolic space.
\item We define a probability distribution in Hyperbolic space for getting noise values and also describe how to sample from the distribution.
\item We evaluate our approach by preserving privacy against an attribution algorithm, baselining against a Euclidean model, and preserving utility on downstream systems.
\end{enumerate}

\section{Privacy Requirement}
Consider a user interacting freely with an \ac{AI} system via a natural language interface. The user's goal is to meet some specific need with respect to an issued query $x$. The expected norm in this specific context would be satisfying the user's request. A privacy violation occurs when $x$ is used to make personal inference beyond what the norm allows \cite{nissenbaum2004privacy}. This generally manifests in the form of unrestricted \ac{PII} present in $x$ (including, but not restricted to locations, medical conditions or personal preferences \cite{schwartz2011pii}). In many cases, the \ac{PII} contains more semantic information than what is required to address the user's base intent and the AI system can handle the request without recourse to the explicit \ac{PII} (we discuss motivating examples shortly). Therefore, our goal is to output $\hat{x}$, a semantic preserving redaction of $x$ that preserves the user's objective while protecting their privacy. We approach this privacy goal along two dimensions (described in Sec.~\ref{sec:privacy_calibration}): (i) uncertainty -- the adversary cannot base their guesses about the user's  \emph{identity} and \emph{property} on information known with certainty from $x$; and (ii) indistinguishability -- the adversary cannot distinguish whether an observed query $\hat{x}$ was generated by a given user's query $x$, or another similar query $x'$.

To describe the privacy requirements and threat model, we defer to the framework provided by \cite{wagner2018technical}. First, we set our \emph{privacy domain} to be the non-interactive \emph{textual} database setting where we seek to release a \emph{sanitized} database to an internal team of \emph{analysts} who can visually inspect the queries. We also restrict this database to the one user, one query model -- i.e., for the baseline, we are not concerned with providing protections on a user's aggregate data. In this model, the analyst is a required part of the system, thus, it is impossible to provide \emph{semantic security} where the analyst learns nothing. This is only possible in a three-party cryptographic system (\eg{} under the Alice, Bob and Eve monikers) where the analyst is different from the attacker (in our threat model, the analyst is simultaneously Bob and Eve). 


We address purely privacy issues by considering that the data was willingly made available by the user in pursuit of a specific objective (as opposed to security issues \cite{bambauer2013privacy} where the user's data might have been stolen). Therefore, we posit that the user's query $x$ is \emph{published} and \emph{observable} data. Our overall aim is to protect the user from \emph{identity} and \emph{property} inference \ie{}, given $x$, the analyst should neither be able to infer with certainty, the user's identity, nor some unique property of the user.

\subsection{Motivating examples}
\label{sec:mech_overview}
To illustrate the desired functionality, which is to infer the user's high level objective while preserving privacy, let us consider the examples in Tab.~\ref{tab:snips_examples} from the Snips dataset \cite{coucke2018snips}:

\begin{table}[h]

\begin{center}
\begin{tabular}{ | l | l | l |}
\hline
 \textbf{Intent} & \textbf{Sample query} & \textbf{New word} \\ 
 \hline
\emph{GetWeather} & will it be colder in \underline{ohio} & (that) state \\
\emph{PlayMusic} & play \underline{techno} on \underline{lastfm}  & music; app \\
\emph{BookRestaurant} & book a restaurant in \underline{milladore}  & (the) city \\
\emph{RateBook} & rate the \underline{firebrand} one of 6 stars  & product \\
\emph{SearchCreativeWork} & i want to watch \underline{manthan} & (a) movie \\

\hline
\end{tabular}
\caption{Examples from the Snips dataset} \label{tab:snips_examples}
\end{center}
\end{table}

\noindent In the examples listed, the underlined terms correspond to the well defined notion of `slot values' while the other words are known as the `carrier phrase'. The slot values are essentially `variables' in queries which can take on different values and are identified by an instance type. We observe therefore that, replacing the slot value with a new word along the similarity or hierarchical axis does not change the user's initial intent. As a result we would expect $\hat{x}$ = `\emph{play (music, song) from app}' to be classified in the same way as $x$ = `\emph{play techno on lastfm}'. We are interested in protecting the privacy of one user, issuing one query, while correctly classifying the user's intent. This model is not designed to handle multiple queries from a user, neither is it designed for handling exact queries \eg{} booking the `specific restaurant in milladore'.

Our objective is to create a privacy preserving mechanism $M$ that can carry out these slot transformations $\hat{x} = M(x)$ in a principled way, with a quantifiable notion of privacy loss.

\section{Privacy mechanism overview}
In this section we review $d_{\chi}$-privacy as a generalization of \ac{DP} over metric spaces. Next, we introduce word embeddings as a natural vector space for $d_{\chi}$-privacy over text. Then, we give an overview of the privacy algorithm in Euclidean space, and finish by discussing why Hyperbolic embeddings would be a better candidate for the privacy task. 

\subsection{Broadening privacy over metric spaces}
Our requirement warrants a privacy metric that confers uncertainty via randomness to an observing adversary while providing indistinguishability on the user inputs and mechanism outputs. Over the years, \ac{DP} \cite{dwork2006calibrating} has been established as mathematically well-founded definition of privacy. It mathematically guarantees that an adversary observing the result of an analysis will make essentially the same inference about any user's information, regardless of whether the user's data is or is not included as an input to the analysis. Formally, \ac{DP} is defined on adjacent datasets $x$ and $x'$ that differ in at most a single row, i.e., the \emph{Hamming distance} between them is at most $1$ and which satisfy the following inequality: 
We say that a randomized mechanism $M : \cX \to \cY$ satisfies $\varepsilon$ \ac{DP} if for any $x, x' \in \cX$ the distributions over outputs of $M(x)$ and $M(x')$ satisfy the following bound: for all $y \in \cY$ we have

\begin{figure}[h]
\begin{subfigure}[t]{\linewidth}
\begin{center}
    \includegraphics[width=.22\linewidth]{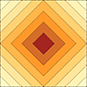}
    \enspace
    \includegraphics[width=.22\linewidth]{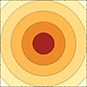}
    \enspace
    \includegraphics[width=.22\linewidth]{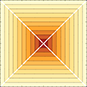}
    \enspace
    \includegraphics[width=.22\linewidth]{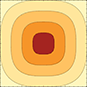}
\end{center}
\end{subfigure}
\caption{Contour plots of different metrics \cite{ruegg_cuda_gael}. Left to right: $L_1$ Manhattan distance, $L_2$ Euclidean distance, $L_\infty$ Chebyshev distance, $L_p$ Minkowski distance ($L_3$ shown here)}
\label{fig:different_metrics}
\end{figure}

{\small
\begin{align}\label{eqn:mdp}
\frac{\Pr[M(x) = y]}{\Pr[M(x') = y]} \leq e^{\varepsilon d(x,x')} \enspace.
\end{align}
}
\noindent where $d(x,x')$ is the Hamming distance and $\varepsilon$ is the measure of privacy loss. \cite{chatzikokolakis2013broadening} generalized the classical definition of \ac{DP} by exploring other possible distance metrics which are suitable where the Hamming distance is unable to capture the notion of closeness between datasets (see Fig.~\ref{fig:different_metrics} for other distance metrics).  

For example, a privacy model built using the Manhattan distance metric can be used to provide indistinguishability when the objective is to release the number of days from a reference point \cite{chatzikokolakis2013broadening}. Similarly, the Euclidean distance on a $2d$ plane can be used to preserve privacy while releasing a user's longitude and latitude to mobile applications \cite{chatzikokolakis2015constructing}. Finally, the Chebyshev distance can be adopted to to perturb the readings of smart meters thereby preserving privacy on what TV channels or movies are being watched \cite{wagner2018technical}.

In order to apply $d_{\chi}$-privacy to the text domain, first, we need a way to organize words in a space equipped with an appropriate distance metric. One way to achieve this is by representing words using a word embedding model.

\subsection{Word embeddings and their metric spaces}
Word embeddings organize discrete words in a continuous metric space such that their similarity in the embedding space reflects their semantic or functional similarity. Word embedding models like Word2Vec \cite{mikolov2013distributed}, GloVe \cite{pennington2014glove}, and fastText\cite{bojanowski2016enriching} create such a mapping $\phi : \cW \to \Reals^n$ of a set of words $\cW$ into n-dimensional Euclidean space. The distance between words is measured by the distance function $d : \cW \times \cW \to \R_+$. This follows as $d(w, w') = d(\phi(w), \phi(w')) = \norm{\phi(w) - \phi(w)}$ where $\norm{\cdot}$ denotes the Euclidean norm on $\R^n$. The vectors $\phi(w_i)$ are generally learned by proposing a conditional probability for observing a word given its context words or by predicting the context giving the original word in a large text corpus \cite{mikolov2013distributed}. 

\subsection{The privacy mechanism in Euclidean space}
Our $d_{\chi}$-privacy algorithm is similar to the model introduced by \cite{wsdm_paper} for privacy preserving text analysis, and \cite{fernandes2018generalised} for author obfuscation. The algorithms are all analogous to that originally proposed by \cite{andres2013geo} and we describe it here using the Euclidean distance for word embedding vectors. In the ensuing sections, we will justify the need to use embedding models trained in Hyperbolic space (Sec.~\ref{sec:case_hyper}) while highlighting the changes required to make the algorithm work in such space. This includes the Hyperbolic word embeddings in Sec.~\ref{sec:pc_embeddings}, describing the noise distribution in Sec.~\ref{sec:hyper_sampling_dist}, and how to sample from it in Sec.~\ref{sec:hyper_sampling_proc}

\begin{algorithm}[h]
\smaller
\DontPrintSemicolon
\SetKw{KwRelease}{release}
\caption{Privacy Mechanism}\label{alg:madlib}
\KwIn{string $x = w_1 w_2 \cdots w_{\ell}$, privacy parameter $\varepsilon > 0$}
\For{$i \in \{1, \ldots, \ell\}$}{
Word embedding $\phi_i = \phi(w_i)$\;
Sample noise $N$ with density $p_N(\vec{z}) \propto \exp(- \varepsilon \norm{\vec{z}})$\;
Perturb embedding with noise $\hat{\phi}_i = \phi_i + N$\;
Discretization $\hat{w}_i = \argmin_{u \in \cW} \norm{\phi(u) - \hat{\phi}_i}$\;
Insert $\hat{w}_i$ in $i$th position of $\hat{x}$\;
}
\KwRelease{$\hat{x}$}
\end{algorithm}

\subsection{The case for Hyperbolic space}
\label{sec:case_hyper}
Even though Euclidean embeddings can model semantic similarity between discrete words in continuous space, they are not well attuned to modeling the latent hierarchical structure of words which are required for our use case. To better capture semantic similarity and hierarchical relationships between words (without exponentially increasing the dimensionality of the embeddings), recent works \cite{nickel2017poincare,nickel2018learning,ganea18a} propose learning the vector representation in Hyperbolic space $\phi : \cW \to \Hyperbolic^n$. Unlike the Euclidean model, the Hyperbolic model can realize word hierarchy through the norms of the word vectors and word similarity through the distance between word vectors (see Eqn.~\ref{eqn:poincare_distance}). Apart from \emph{hypernymy} relationships (e.g., {\footnotesize \textsc{London} $\to$ \textsc{England}}), Hyperbolic embeddings can also model multiple latent hierarchies for a given word (e.g., {\footnotesize \textsc{London} $\to$ \textsc{Location}} and {\footnotesize \textsc{London} $\to$ \textsc{City}}). Capturing these \textsc{is-a} relationships (or concept hierarchies) using Hyperbolic embeddings was recently demonstrated by \cite{le2019inferring} using data from large text corpora. 

Furthermore, for Euclidean models such as \cite{wsdm_paper,fernandes2018generalised}, the utility degrades badly as the privacy guarantees increase. This is because the noise injected (line $4$ of Alg.~\ref{alg:madlib}) increases to match the privacy guarantees, resulting in words that are not semantically related to the initial word. The space defined by Hyperbolic geometry (Sec~\ref{sec:hyp_space_geo}), in addition to the distribution of words as concept hierarchies does away with this problem while preserving privacy and utility of the user's query.

\section{Hyperbolic space and Geometry}
\label{sec:hyp_space_geo}
Hyperbolic space $\Hyperbolic^n$ is a homogeneous space with constant negative curvature \cite{krioukov2010hyperbolic}. The space exhibits hyperbolic geometry, characterized by a negation of the parallel postulate with infinite parallel line passing through a point. It is thus distinguished from the other two isotropic spaces: Euclidean $\Reals^n$, with zero (flat) curvature; and spherical $\Spherical^n$, with constant positive curvature. Hyperbolic spaces cannot be embedded isometrically into Euclidean space, therefore embedding results in every point being a \textit{saddle point}. In addition, the growth of the hyperbolic space area is exponential (with respect to the curvature $K$ and radius $r$), while Euclidean space grows polynomially (see Tab.~\ref{tab:euc_hyp} for a summary of both spaces). 

\begin{table}[h]
\smaller
\begin{center}
\begin{tabular}{ | l | l | l | }
 \hline
 \textbf{Property} & \textbf{Euclidean} & \textbf{Hyperbolic} \\ 
 \hline
 Curvature $K$ & $0$ & $< 0$ \\  
 \hline
 Parallel lines & $1$ & $\infty$ \\
 \hline
 Triangles are & normal & thin \\
 \hline
 Sum of $\bigtriangleup$ angles & $\pi$ & $< \pi $ \\
 \hline
 Circle length & $2 \pi r$ & $2 \pi$ sinh $\zeta r $ \\
 \hline
 Disk area & $2 \pi r^2 / 2$ & $2 \pi$ (cosh $\zeta r - 1)$ \\
 \hline
\end{tabular}
\caption{Properties of Euclidean and hyperbolic geometries. Parallel lines is the number of lines parallel to a line and that go through a point not on this line, and $\zeta =  \sqrt{|K|}$ \cite{krioukov2010hyperbolic}} \label{tab:euc_hyp}
\end{center}
\end{table}

As a result of the unique characteristics of hyperbolic space, it can be constructed with different isomorphic models. These include: the Klein model, the \pc{} disk model, the \pc{} half-plane model, and the Lorentz (or hyperboloid) model. In this paper, we review two of the models: the Lorentz model, and the \pc{} model. We also highlight what unique properties we leverage in each model and how we can carry out transformations across them.

\subsection{\pc{} ball model}
\begin{figure*}
  \begin{subfigure}[b]{0.30\textwidth}
    \includegraphics[width=\textwidth]{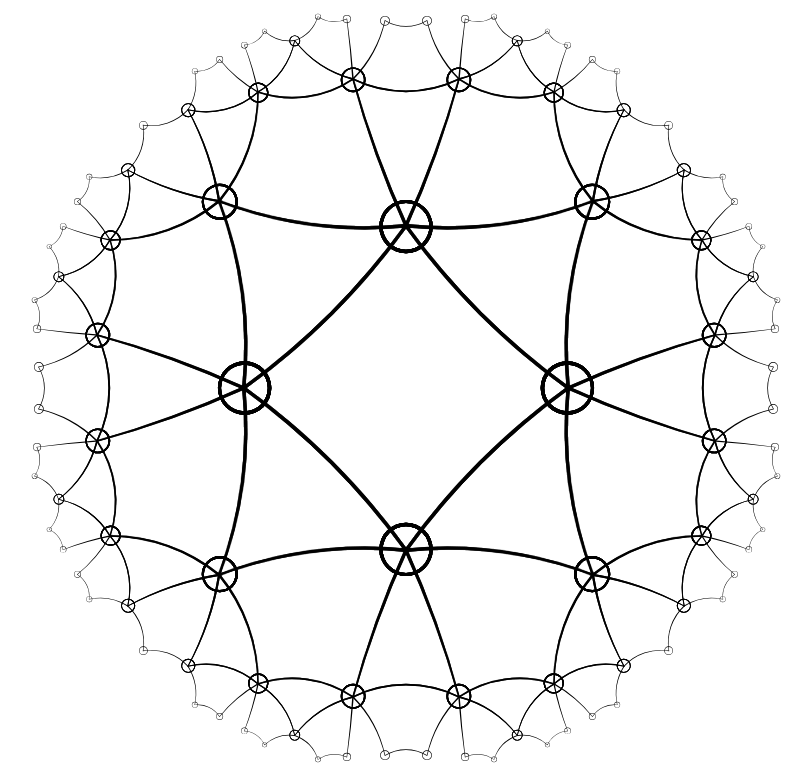}
    \caption{}
    \label{fig:model_a}
\end{subfigure}
  \begin{subfigure}[b]{0.35\textwidth}
    \includegraphics[width=\textwidth]{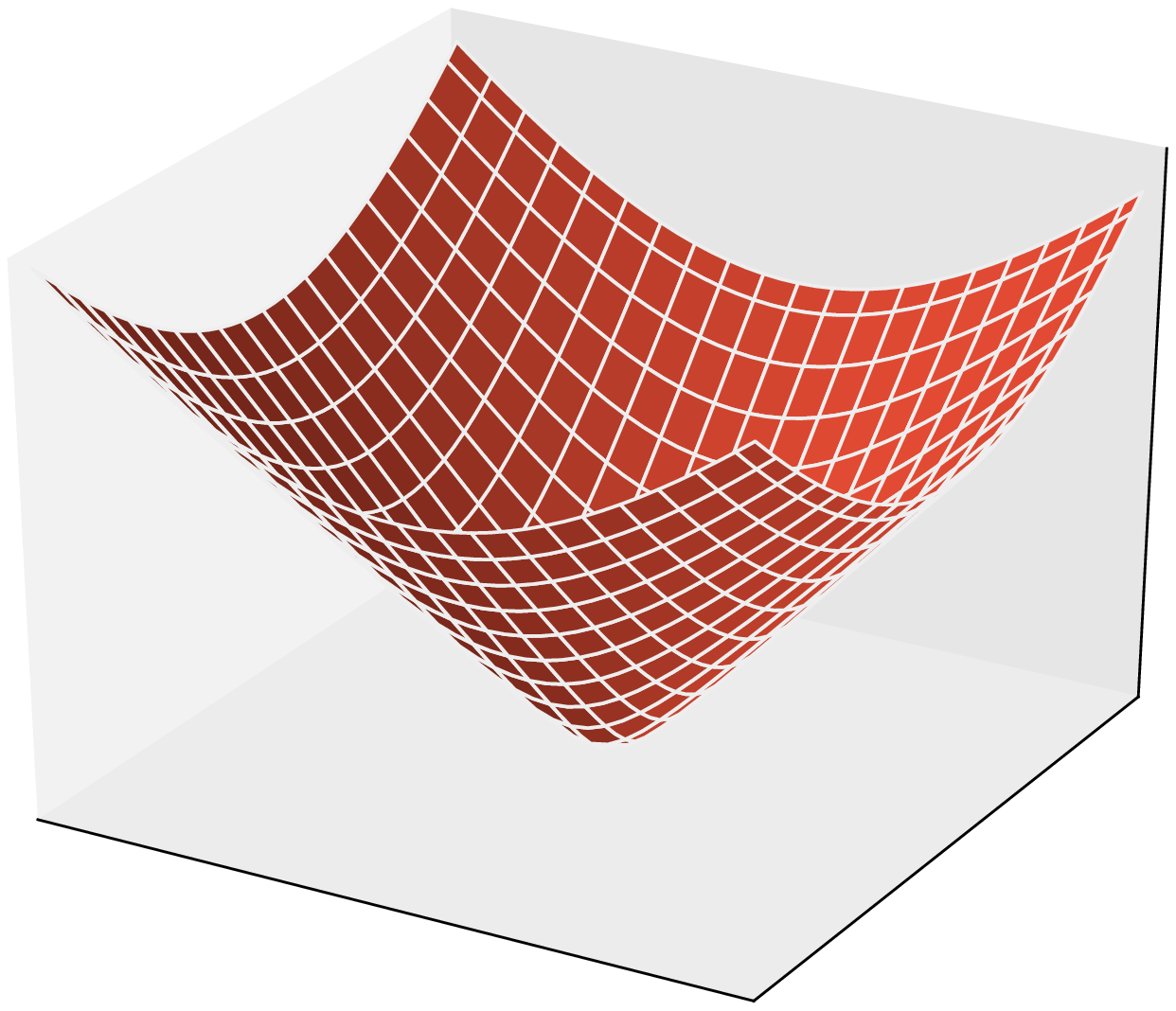}
    \caption{}
    \label{fig:model_b}
\end{subfigure}
  \begin{subfigure}[b]{0.32\textwidth}
    \includegraphics[width=\textwidth]{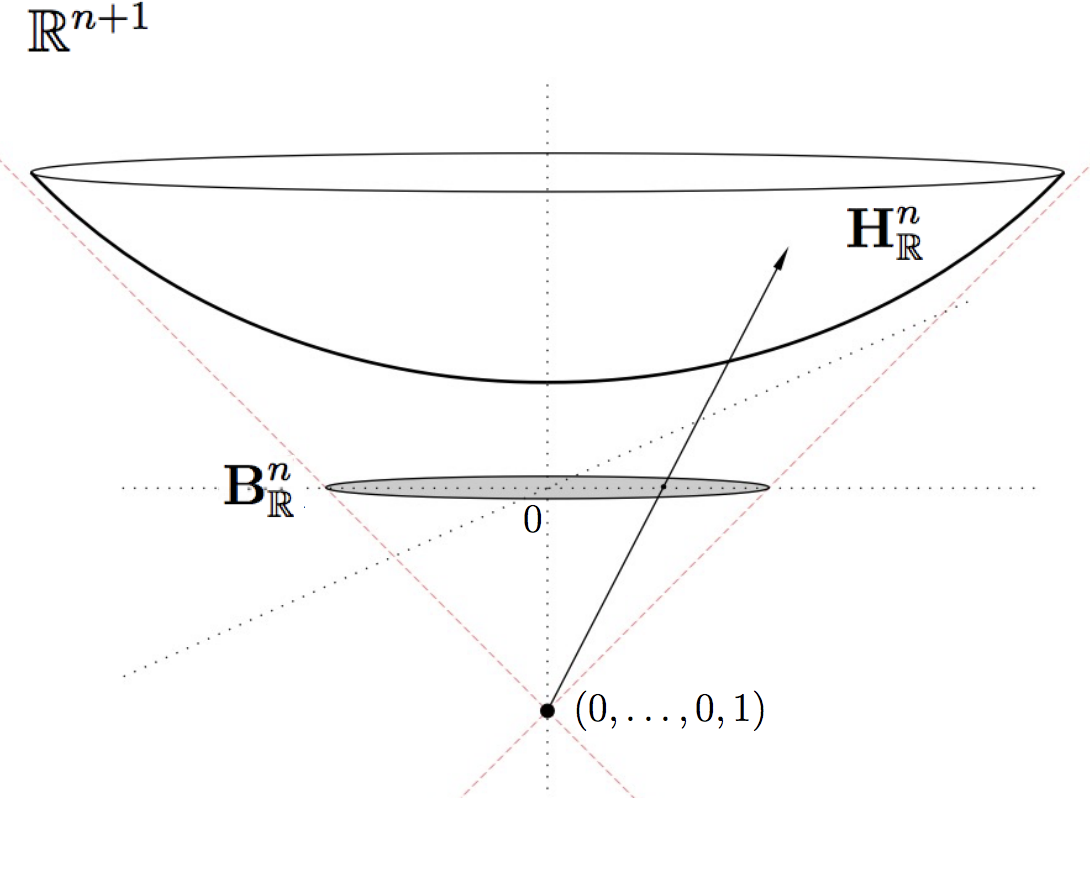}
    \caption{}
    \label{fig:model_c}
\end{subfigure}

    \caption{(a) Tiling a square and triangle in the \pc{} disk $\mathcal{B}^2$ such that  all line segments have identical hyperbolic length. (b) The forward sheet of a two-sheeted hyperboloid in the Lorentz model. (c) Projection \cite{staleymodels} of a point in the Lorentz model $\mathcal{H}^n$ to the \pc{} model $\mathcal{B}^n$ (d) Embedding WebIsADb \textsc{is-a} relationships in the GloVe vocabulary into the $\mathcal{B}^2$ \pc{} disk}
\label{fig:hyperbolic_models}
\end{figure*}

The $n-$dimensional \pc{} ball $\mathcal{B}^n$ is a model of hyperbolic space $\Hyperbolic^n$ where all the points are mapped within the $n-$dimensional open unit ball i.e., $\mathcal{B}^n = \{ x \in \Reals^n | \LN x \RN < 1 \}$ where $\LN \cdot \RN$ is the Euclidean norm. The boundary of the ball i.e.,  the hypersphere $\Spherical^{n-1}$ is not part of the hyperbolic space, but it represents points that are infinitely far away (see Fig.~\ref{fig:model_a}). The \pc{} ball is a conformal model of hyperbolic space (i.e., Euclidean angles between hyperbolic lines in the model are equal to their hyperbolic values) with metric tensor: $g_p(x) = [2 / (1 - \LN x \RN^2)]^2 \enspace g_e$ 
 \footnote{The metric tensor (like a dot product) gives \textit{local} notions of length and angle between tangent vectors. By integration local segments, the metric tensor allows us to calculate the \textit{global} length of curves in the manifold} 
where $x \in \mathcal{B}^n$ and $g_e$ is the Euclidean metric tensor. The \pc{} ball model then corresponds to the Riemannian manifold $\mathcal{P}^n = (\mathcal{B}^n, g_p)$. Considering that the unit ball represents the infinite hyperbolic space, we introduce a distance metric
by: $d \rho = 2dr / (1 - r^2)$
where $\rho$ is the \pc{} distance and $r$ is the Euclidean distance from the origin. Consequently, the growth in distance $d \rho \to \infty$ as $r \to 1$, which proves the infinite extent of the ball. Therefore, given $2$ points (\eg{} representing word vectors) $u, v \in \mathcal{B}^n$ we define the isometric invariant: \cite{staleymodels} 
{\small
\begin{align*}
\delta(u,v) = 2 \frac{ \LN u - v \RN^2 } { (1 - \LN u \RN^2) (1 - \LN v \RN^2) }
\end{align*}
}

\noindent then the distance function over $\mathcal{P}^n$ is given by:
{\small
\begin{align*}
d(u,v) &= \arcosh(1 + \delta(u,v)) \\
    	  &= \arcosh \LP 1 + 2 \frac{ \LN u - v \RN^2 } { (1 - \LN u \RN^2) (1 - \LN v \RN^2) } \RP \numberthis \label{eqn:poincare_distance}
\end{align*}
}

The main advantage of this model is that it is conformal - as a result, earlier research into Hyperbolic word embeddings have leveraged on this model \cite{alanis2016efficient,ganea18a,tifrea2018poincar}. Furthermore, there were existing artifacts such as the \pc{} embeddings by \cite{nickel2017poincare} built with this model that we could reuse for this work.

\subsection{Lorentz model}
The Lorentz model (also known as the hyperboloid or Minkowski model) is a model of hyperbolic space $\Hyperbolic^n$ in which points are represented by the points on the surface of the upper sheet of a two-sheeted hyperboloid in ($n+1$)-dimensional Minkowski space. It is a combination of $n-$dimensional spatial Euclidean coordinates $x^k_i$ for $k = 1, 2, ..., n$; and a time coordinate $x^0_i > 0$. Therefore, given points $u,v \in \Reals^{n+1}$, the Lorentzian inner product (Minkowski bilinear form) is:

{\small
\begin{align}
\label{eqn:lorentz_scalar_prod}
\LA u, v \RA_\mathcal{L} = -u_0 v_0 + \sum^n_{i=1} u_i v_i
\end{align}
}
\noindent The product of a point with itself is $-1$, thus, we can compute the norm as $\LN x \RN_\mathcal{L} =  \sqrt{\LA x, x \RA_\mathcal{L}}$. We define the Lorentz model as a Riemannian manifold $\mathcal{L}^n = (\mathcal{H}^n, g_l)$ where: $\mathcal{H}^n = \{ u \in \Reals^{n+1} : \LA u, v \RA_\mathcal{L} = -1, u_0 > 0 \}$
and the metric tensor $g_l = diag(+1, -1, ..., -1)$. Hence, given the vector representation of a word at the origin in Euclidean space $\Reals^n$ as $[ 0, 0, ... 0 ]$, the word's corresponding vector location in the Hyperboloid model $\Reals^{n+1}$ is $[1, 0, ..., 0]$ where the first coordinate $x_0$ for $x = (x_0, x') \in \Reals^{n+1}$ is:

{\small
 \begin{align}
\label{eq:x0_point}
x_0 \in \mathcal{H}^n = \sqrt{1 + \LN x' \RN^2} \enspace \text{where} \enspace x' = (x_1, ..., x_n)
\end{align}
}

The hyperbolic distance function admits a simple expression in $\mathcal{L}^n$ and it is given as:
{\small
 \begin{align}
d_l(u,v) = \arcosh \LP - \LA u, v \RA_\Lcal \RP
\end{align}
}
This distance function satisfies the axioms of a metric space (\ie{} identity of indiscernibles, symmetry and the triangle inequality). Its simplicity and satisfaction of the axioms make it the ideal model for constructing our privacy proof.

\subsection{Connection between the models}
\label{sec:hyper_equiv}
Both models essentially describe the same structure of hyperbolic space characterized by its constant negative curvature. They simply represent different coordinate charts in the same metric space. Therefore, the Lorentz and \pc{} model can be related by a diffeomorphic transformation that preserves all the properties of the geometric space, including isometry. From Fig.~\ref{fig:model_c}, we observe that a point $x_{\mathcal{P}}$ in the \pc{} model is a projection from the point $x_{\mathcal{L}} = (x_0, x')$ in the Lorentz model, to the hyperplane $x_0 = 0$ by intersecting it with a line drawn through $[-1, 0, ..., 0]$. Consequently, we can map this point across manifolds from the Lorentz to the \pc{} model via the transformation $x_{\mathcal{P}} : \mathcal{L}^n \to \mathcal{P}^n$ where:

{\small
\begin{align}
\label{eq:lorentz_to_poincare}
x_{\mathcal{P}} = \frac{x'}{1 + x_0}  \enspace \text{where} \enspace x' = (x_1, ..., x_n)
\end{align}
}

In this work, we only require transformations to the \pc{} model i.e., using Eqn.~\ref{eq:x0_point} and \ref{eq:lorentz_to_poincare}. Mapping points back from \pc{} to Lorentz is done via:
\begin{align*}
x_{\mathcal{L}} = (x_0, x') = \frac{(1 + \LN x \RN^2, 2x)}{1 - \LN x \RN^2}
\end{align*}
As a result of the equivalence of the models, in this paper, we adopt the Lorentz model for constructing our $d_{\chi}$-privacy proof while the word embeddings were trained in the \pc{} ball model. Consequently, the \pc{} model is also used as the basis for sampling noise from a high dimensional distribution to provide the privacy and semantic guarantees.

\section{Privacy proof and Sampling mechanism}
In this section, we provide the proof of $d_{\chi}$-privacy for Hyperbolic space embeddings. We will be using the Lorentz model of \cite{nickel2018learning} rather than the \pc{} model proposed in \cite{nickel2017poincare}. Then in Sec.~\ref{sec:hyper_sampling_dist}, we introduce our probability distribution for adding noise (line $4$ of Alg.~\ref{alg:madlib}) to the word embedding vectors. Finally, we describe how to sample (line $3$ of Alg.~\ref{alg:madlib}) from the proposed distribution in Sec.~\ref{sec:hyper_sampling_proc}. We note that whereas the privacy proof is provided in the Lorentz model, the noise distribution and the embeddings are in the \pc{} model. See Sec.~\ref{sec:hyper_equiv} for discussions on the equivalence of the models.

\subsection{$d_{\chi}$-privacy proof}
\label{sec:hyper_proof}
In this section, we will show $d_{\chi}$-privacy for the Hyperboloid embeddings of \cite{nickel2018learning}.  In the following, given $u, v \in \Reals^{n + 1}$, we use the Lorentzian inner product from Eqn~\ref{eqn:lorentz_scalar_prod} \ie{} $\LA u, v \RA_\Lcal = -u_0 v_0 + \sum_{i=1}^n u_i v_i.$
%
%
The space $(\Hyperbolic^n, d)$, where $d(u, v) = \arcosh \LP - \LA u, v \RA_\Lcal \RP \in [0, \infty] ,$ is the hyperboloid model of $n$-dimensional (real) hyperbolic space.

\begin{lemma} 
If $u, v \in \Hyperbolic^n$, then $\LA u, v \RA_\Lcal \leq -1$ with equality only if $u = v$.
\begin{proof}
Using the Cauchy-Schwarz inequality for the Euclidean inner product in $\Reals^n$ for the first inequality and a simple calculation for the second, we have:
{\small
\begin{align*} 
\Bigg(\sum_{i = 1}^ n u_i v_i\Bigg)^2 &= \Bigg(\sum_{i = 1}^n u_ i^ 2\Bigg) \Bigg(\sum_{i = 1}^n v_ i^ 2\Bigg) \enspace \text{from Cauchy-Schwarz, then}\\
\LA u, v \RA_\Lcal &= -u_0 v_0 + \sum_{i = 1}^ n u_i v_i \leq - u_0 v_0 + \sqrt{\sum_{i = 1}^n u_ i^ 2} \sqrt{\sum_{i=1}^n v_i^2 } \\ 
                              &= -u_0 v_0 + \sqrt{u_0^2 - 1} \sqrt{v_0^2 - 1} \leq - 1 
\end{align*}
}
Any line through the origin intersects $\Hyperbolic^n$ in at most one point, so Cauchy's inequality is an equality if and only if $u = v$ (as a consequence of using the positive roots).

\end{proof}
\end{lemma}

Now, as was the case for the Euclidean metric, we can use the triangle inequality for the metric $d$ which implies that for any $z \in \Reals^n$ we have the following inequality:

{\small
\begin{align*}
\exp &\LP -\varepsilon d(z, \phi(w)) \RP = \frac{ \exp \LP -\varepsilon d(z, \phi(w)) \RP } { \exp \LP -\varepsilon d(z, \phi(w')) \RP} \exp \LP -\varepsilon d(z, \phi(w')) \RP \\
&= \exp \Big( \varepsilon d(z, \phi(w')) -\varepsilon d(z, \phi(w)) \Big) \times \exp \Big( -\varepsilon d(z, \phi(w')) \Big) \\
&= \exp \Bigg[ \varepsilon \arcosh \LP - \LA z, \phi(w') \RA_\Lcal \RP - \varepsilon \arcosh \Big( - \LA z, \phi(w) \RA_\Lcal \Big) \Bigg] \\
&\times \exp \LP -\varepsilon d(z, \phi(w') \RP \\
&<= \exp \Bigg( \varepsilon \arcosh \LP \LA \phi(w), \phi(w') \RA_\Lcal \RP \Bigg) \exp \Big( -\varepsilon d(z, \phi(w') \Big) \\
&= \exp \Big( \varepsilon d \LP \phi(w), \phi(w') \RP \Big) \exp \Big( -\varepsilon d(z, \phi(w') \Big) \\
\end{align*}
}
Thus, as before by plugging the last two derivations together and observing the the normalization constants in $p_N(z)$ and $p_{\phi(w) + N}(z)$ are the same, we obtain:

{\small
\begin{align*}
\frac {\Pr [M(w) = \hat{w} ] } { \Pr \LB M \LP w' \RP = \hat{w} \RB} &= \frac{\int_{C_{\hat{w}}} \exp(-\varepsilon \LA z, \phi(w) \RA_\Lcal) dz} { \int_{C_{\hat {w}}} \exp \left( - \varepsilon \LA z, \phi(w') \RA_\Lcal \RP dz } \\
& \leq \exp \left( \varepsilon d \left( w , w ^ { \prime } \right) \right)
\end{align*}
}
Thus, for $l = 1$ the mechanism $M$ is $\varepsilon d_{\chi}$-privacy preserving. The proof for $l > 1$ is identical to the Euclidean case of \cite{wsdm_paper}.

\subsection{Probability distribution for sampling noise}
\label{sec:hyper_sampling_dist}
In this section we describe the Hyperbolic distribution from which we sample our noise perturbations. One option was sampling from the Hyperbolic normal distribution proposed by \cite{said2014new} (discussed in \cite{mathieu2019hierarchical} and \cite{ovinnikov2019poincar}) with pdf:

{\small
\begin{align*}
	p(x | \mu, \sigma) = \frac{1}{Z(\sigma)} e^{\frac{d^2(x, \mu)}{2\sigma^2}} \enspace  \text{and} \enspace Z(\sigma) = 2 \pi \sqrt{\frac{\pi}{2}} \sigma e^{\frac{\sigma^2}{2}} \text{erf} \Bigg(\frac{\sigma}{\sqrt{2}} \Bigg)
\end{align*}
}

\noindent However, our aim was to sample from the family of Generalized Hyperbolic distributions which reduce to the Laplacian distribution at particular location and scale parameters. By taking this approach, we can build on the proof proposed in the Euclidean case of $d_{\chi}$-privacy where noise was sampled from a planar Laplace distribution \cite{andres2013geo,chatzikokolakis2013broadening}.

In the \pc{} model of Hyperbolic spaces, we have the following distance function defined in Eqn~\ref{eqn:poincare_distance}:
{\small
\begin{align*}
    d(u, v) = \arcosh \LP 1 + 2 \frac{ \LN u - v \RN^2 } { (1 - \LN u \RN^2) (1 - \LN v \RN^2) } \RP
\end{align*}
}
Now, analogous to the Euclidean distance used for the Laplace distribution, we wish to construct a distribution that matches this distance function. This will take the form:
{\small
\begin{align*}
    p(x | \mu, \varepsilon) &\propto \LP -\varepsilon d(x, \mu) \RP \\
                                       &= \exp \LP -\varepsilon \arcosh \LP 1 + 2 \frac{ \LN x - \mu \RN^2 } { (1 - \LN x \RN^2) (1 - \LN \mu \RN^2) } \RP \RP
\end{align*}
}
In all cases, our noise will be centered at $x$, and hence $\mu = 0$:
{\small
\begin{align*}
    p(x | \mu=0, \varepsilon) &= \frac{1}{Z} \exp \LP -\varepsilon d(x, 0) \RP \\
                                           &= \frac{1}{Z} \exp \Bigg[ -\varepsilon \arcosh \LP 1 + 2 \frac{ \LN x \RN^2 } { (1 - \LN x \RN^2)  } \RP \Bigg] 
\end{align*}
}

\noindent Next, we set a new variable $c$ and observe:
{\small
\begin{align*}
c =  \frac{ \LN x \RN^2 } { (1 - \LN x \RN^2)  }  \enspace and \enspace \arcosh(1 + 2c) = \log \Big( 2c + 2 \sqrt{c^2 + c} + 1 \Big)
\end{align*}
}

{\small
\begin{align*}
    p(x | \mu=0, \varepsilon) &= \frac{1}{Z} \exp \Bigg[ -\varepsilon \log \LP 1 + 2 c + \sqrt{ \LP 1 + 2 c  \RP^2 - 1 } \RP  \Bigg] 
\end{align*}
}

\noindent Reinserting the variable $c$ and simplifying
{\small
\begin{align*}
                                           &= \frac{1}{Z} \exp \Bigg[ -\varepsilon \log \LP -\frac{2}{ \LN x \RN - 1} - 1 \RP \Bigg] \\
                                           &= \frac{1}{Z} \LP - \frac{2}{\LN x \RN - 1} -1 \RP^{-\varepsilon}
\end{align*}
}

\begin{figure}[h]
    \includegraphics[width=0.9\linewidth]{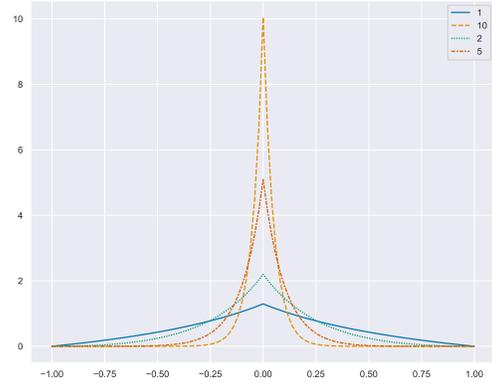}
    \caption{$1d$ PDF of Eqn.~\ref{eq:pdf} at different values of $\varepsilon$}
    \label{fig:laplace_pdf}
  \end{figure}

\noindent The normalization constant is then:
{\small
\begin{align*}
	Z &= \int_{-1}^{1} \LP - \frac{2}{\LN x \RN - 1} -1 \RP^{-\varepsilon} dx \\
	   &= \frac{2 \; \hg(1, \varepsilon, 2 + \varepsilon, -1)}{1 + \varepsilon}
\end{align*}
}
where $\hg(a, b; c, z)$ is the hypergeometric function defined for $|z| < 1$ by the power series
{\small
\begin{align*}
	\hg(a,b;c;z)=\sum _{n=0}^{\infty }{\frac {(a)_{n}(b)_{n}}{(c)_{n}}}{\frac {z^{n}}{n!}}.
\end{align*}
}
where $(z)_n = \frac{\Gamma(z + n)}{\Gamma(z)}$ is the Pochhammer symbol (rising factorial). 
Note that $Z$ does not depend on $x$, and hence can be computed in closed form $\emph{a-priori}$.
Our distribution is therefore:
{\small
\begin{align}
\label{eq:pdf}
p(x | \mu=0, \varepsilon) = \frac{1 + \varepsilon}{2 \; \hg(1, \varepsilon, 2 + \varepsilon, -1)} \LP - \frac{2}{\LN x \RN - 1} -1 \RP^{-\varepsilon}
\end{align}
}
The result shown in Fig.~\ref{fig:laplace_pdf} for different values of $\varepsilon$ illustrates the PDF of the new distribution derived from Eqn.~\ref{eq:pdf}.

\subsection{Sampling from the distribution}
\label{sec:hyper_sampling_proc}
Since we are unable to sample directly from the high dimensional hyperbolic distribution in Eqn.~\ref{eq:pdf}, we derive sampled points by simulating random walks over it using the \ac{MH} algorithm. A similar approach was adopted by \cite{mathieu2019hierarchical} and \cite{ovinnikov2019poincar} to sample points from high dimensional Riemannian normal distributions using Monte Carlo samples. We start with $f(x, \varepsilon)$ which is our desired probability distribution as defined in Eqn.~\ref{eq:pdf} where $\varepsilon$ is the $d_{\chi}$-privacy parameter. Then we choose a starting point $x_0$ to be the first sample. The point $x_0$ is set at the origin of the \pc{} model (see Fig.~\ref{fig:model_c}). The sample $x_0$ is then updated as the current point $x_t$.

To select the next candidate $x_t'$, \ac{MH} requires the point be sampled ideally from a symmetric distribution $g$ such that $g(x_t | x_t') = g(x_t' | x_t)$ for example, a Gaussian distribution centered at $x_t$. 
To achieve this, we sampled $x_t'$ from the multivariate normal distribution in Euclidean space, centered at $x_t$. The sampled point $x_t$ is then translated to the $\mathcal{H}^n$ Lorentz model in $\Reals^{n+1}$ dimensional Minkowski space by setting the first coordinate using Eqn.~\ref{eq:x0_point}. The Lorentz coordinates are then converted to the $\mathcal{B}^n$ \pc{} model in $\Reals^n$ Hyperbolic space using Eqn.~\ref{eq:lorentz_to_poincare}. Therefore, the final coordinates of the sampled point $x_t$ is in the \pc{} model.


Next, for every \ac{MH} iteration, we calculate an acceptance ratio $\alpha = f(x_t', \varepsilon)/f(x_t, \varepsilon)$ with our privacy parameter $\varepsilon$. If $\alpha$ is less than a uniform random number $u \sim \mathcal{U}([0,1])$, we accept the sampled point by setting $x_{t+1} = x_t'$ (and sample the next point centered at this new point), otherwise, we reject the sampled point by setting $x_{t+1}$ to the old point $x_t$. 


\begin{algorithm}[h]
\smaller
\DontPrintSemicolon
\SetKw{KwRelease}{release}
\caption{Hyperbolic Noise Sampling Mechanism}\label{alg:hyper_madlib}
\KwIn{dimension $n > 0$, $\mu = 0$, privacy parameter $\varepsilon > 0$}
\KwResult{$k$ results from $\mathcal{B}^n$}
Let $f(x, \varepsilon)$ be the Hyperbolic noise distribution in $n$ dimensions\;
set $x_0 = [1, 0, ..., 0]$ \;
set $x_t = x_0$\;
set $b$ as the initial sample burn-in period\;
\While{$i < k + b $}{
sample $x' \sim \mathcal{N}(x_t, \Sigma) $\;
translate $x' \to \mathcal{H}^n \to \mathcal{B}^n$\;
compute $\alpha = f(x') / f(x_t)$\;
sample $u \sim \mathcal{U}[(0, 1)]$\;
\uIf{$u \leq \alpha $}{
    accept sample \;
    set $x_{t+1} = x'$ \;
  }
  \Else{
    reject sample \;
    set $x_{t+1} = x_t$ \;
  }
}
\KwRelease{$x^n_i, ..., x^n_k$}
\end{algorithm}

\subsection{Ensuring numerical stability}
Sampling in high dimensional Hyperbolic spaces comes with numeric stability issues \cite{nickel2017poincare,nickel2018learning,mathieu2019hierarchical}. This occurs as the curvature and dimensionality of the space increases. This leads to points being consistently sampled at an infinite distance from the mean. Using an approach similar to \cite{nickel2017poincare}, we constrain the updated vector to remain within the \pc{} ball by updating the noisy vectors as:
{\small
\[
    proj(\theta) \; = \; 
\begin{dcases}
    \theta / \LN \theta \RN \cdot (1 + \lambda), & \text{if} \LN \theta \RN \geq 1\\
    \theta,              & \text{otherwise}
\end{dcases}
\]
}

\noindent where $\lambda$ is a small constant. This occurs as a post-processing step and therefore does not affect the $d_{\chi}$-privacy proof. In our experiments, we set the value to be $\lambda = 10e{-5}$ as in \cite{nickel2017poincare}.


\subsection{\pc{} embeddings for our analysis}
\label{sec:pc_embeddings}
\begin{figure}[h]
    \includegraphics[width=0.9\linewidth]{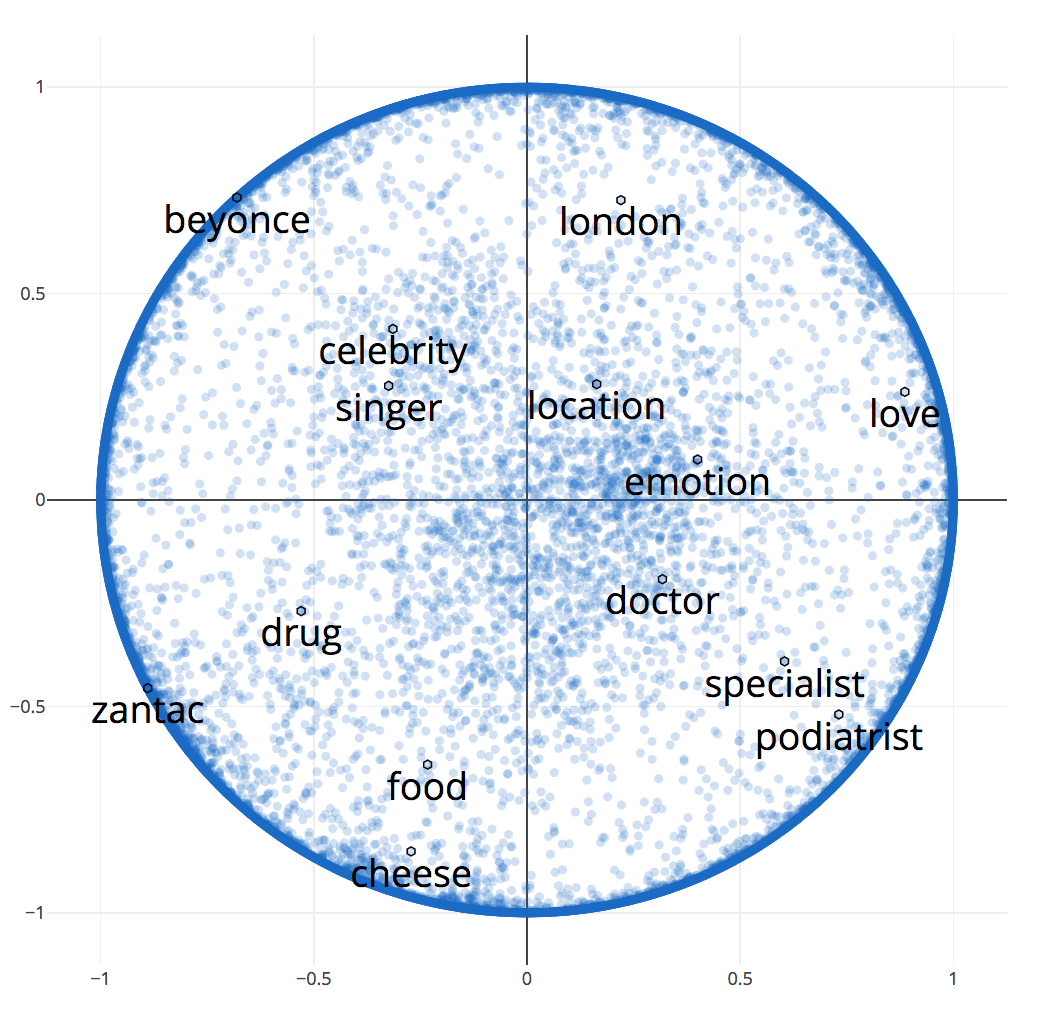}
    \caption{Embedding WebIsADb \textsc{is-a} relationships in the GloVe vocabulary into the $\mathcal{B}^2$ \pc{} disk}
    \label{fig:poincare_disk_results}
  \end{figure}

The geometric structure of the \pc{} embeddings represent the metric space over which we provide privacy guarantees. 
By visualizing the embeddings in the \pc{} disk (see Fig.~\ref{fig:poincare_disk_results}), we observe that higher order concepts are distributed towards the center of the disk, instances are found closer to the perimeter, and similar words are equidistant from the origin.

In this work, we train the \pc{} embeddings described in \cite{nickel2017poincare}. To train, we use data from \textsc{WebIsADB}, a large database of over $400$ million hypernymy relations extracted from the CommonCrawl web corpus. We narrowed the dataset by only selecting relations of words (i.e., both the instance and the class) that occurred in the \textsc{GloVe} vocabulary. To ensure we had high quality data, we further restricted the data to links that had been found at least $10$ times in the CommonCrawl corpus. Finally, we filtered out stop words, offensive words and outliers (words with $\leq 2$ links) from the dataset, resulting in $\approx126,000$ extracted \textsc{is-a} relations. We use the final dataset to train a $100-$dimension \pc{} embedding model. We use this model for all our analysis and experiments.

\section{Privacy calibration}
\label{sec:privacy_calibration}
In this section, we describe our approach for calibrating the values of $\varepsilon$ for a given mechanism $M$. 
For all our discussions, $M(w) = w$ means the privacy mechanism $M$ returns the same word, while $M(w) = \hat{w}$ represents a different random word from the algorithm. 
We now define the privacy guarantees that result in \emph{uncertainty} for the adversary over the \emph{outputs} of $M(w)$, and \emph{indistinguishability} over the \emph{inputs} to $M(w)$.

\subsection{Uncertainty statistics}
The uncertainty of an adversary is defined over the probability of predicting the value of the random variable $\hat{w}$ \ie{} $\Pr[M(w) = \hat{w}]$. This follows from the definition of \emph{Shannon entropy} which is the number of additional bits required by the adversary to reveal the user's \emph{identity} or some secret \emph{property}. However, even though entropy is a measure of uncertainty, there are issues with directly adopting it as a privacy metric \cite{wagner2018technical} since it is possible to construct different probability distributions with the same level of entropy. 

Nevertheless, we still resort to defining the uncertainty statistics by using the two extremes of the R{\'e}nyi entropy \cite{renyi1961measures}. The Hartley entropy $H_0$ is the special case of R{\'e}nyi entropy with $\alpha = 0$. It depends on vocabulary size $|\cW|$ and is therefore a best-case scenario as it represents the perfect privacy scenario for a user as the number of words grow. It is given by $H_0 = \log_2 |\cW| $. Min-entropy $H_{\infty}$ is the special case with $\alpha = \infty$ which is a worst-case scenario because it depends on the adversary attaching the highest probability to a specific word $p(w)$. It is given by $H_{\infty} = -\log_2 \underset{w \in \cW}{\max}(p(w))$.

We now describe proxies for the Hartley and Min-entropy. First, we observe that the mechanism $M(w)$ at $\varepsilon = (0, \infty)$ has full support over the entire vocabulary $\cW$. However, empirically, the effective number of new words returned by the mechanism $M(w)$ over multiple runs approaches a finite subset. As a result, we can expect that $|\hat{\cW}|_{\varepsilon \to 0} > |\hat{\cW}|_{\varepsilon \to \infty}$ for a finite number of successive runs of the mechanism $M(w)$. We define this effective number $|\hat{\cW}|$ at each value of $\varepsilon$ for each word as $S_w$. Therefore, our estimate of the Hartley entropy $H_0$ becomes $H_0 = \log_2 |\cW| \approx \log_2 S_w$. 

Similarly, we expect that over multiple runs of the mechanism $M(w)$, as $\varepsilon \to \infty$, the probability $\Pr[M(w) = w]$ increases and approaches $1$. As a result, we can expect that $\Pr[M(w) = w]_{\varepsilon \to 0} < \Pr[M(w) = w]_{\varepsilon \to \infty}$ for a finite number of successive runs of the mechanism $M(w)$. We define this number $\Pr[M(w) = w]$ at each value of $\varepsilon$, and for each word as $N_w$. Therefore, our estimate of the Min-entropy $H_\infty$ becomes $H_{\infty} = -\log_2 \underset{w \in \cW}{\max}(p(w)) \approx -\log_2 N_w$. 

\begin{figure*}
  \begin{subfigure}[b]{0.245\textwidth}
    \includegraphics[width=\textwidth]{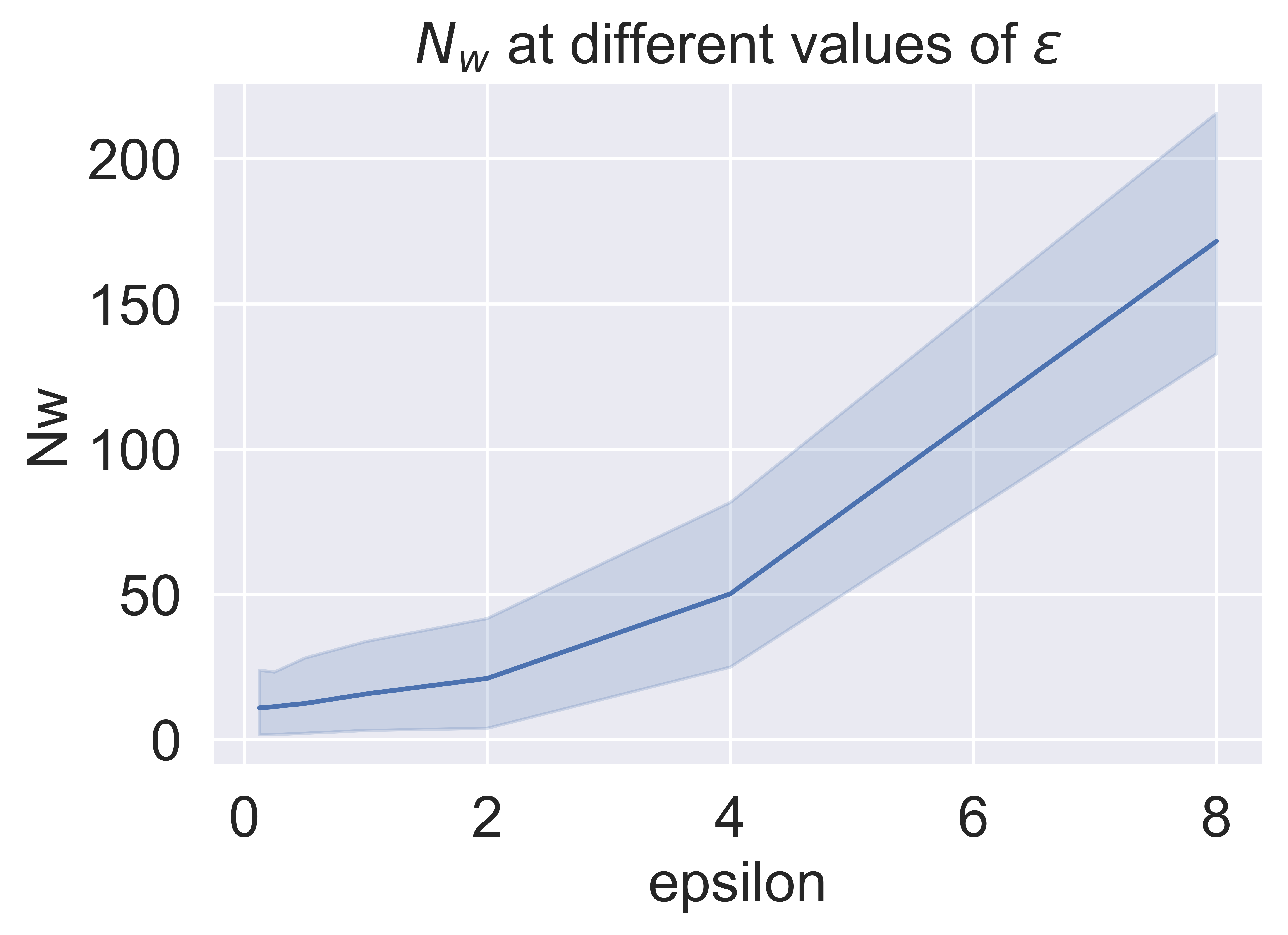}
    \caption{$N_w$ average case}
    \label{fig:nw_stats}
\end{subfigure}
  \begin{subfigure}[b]{0.25\textwidth}
    \includegraphics[width=\textwidth]{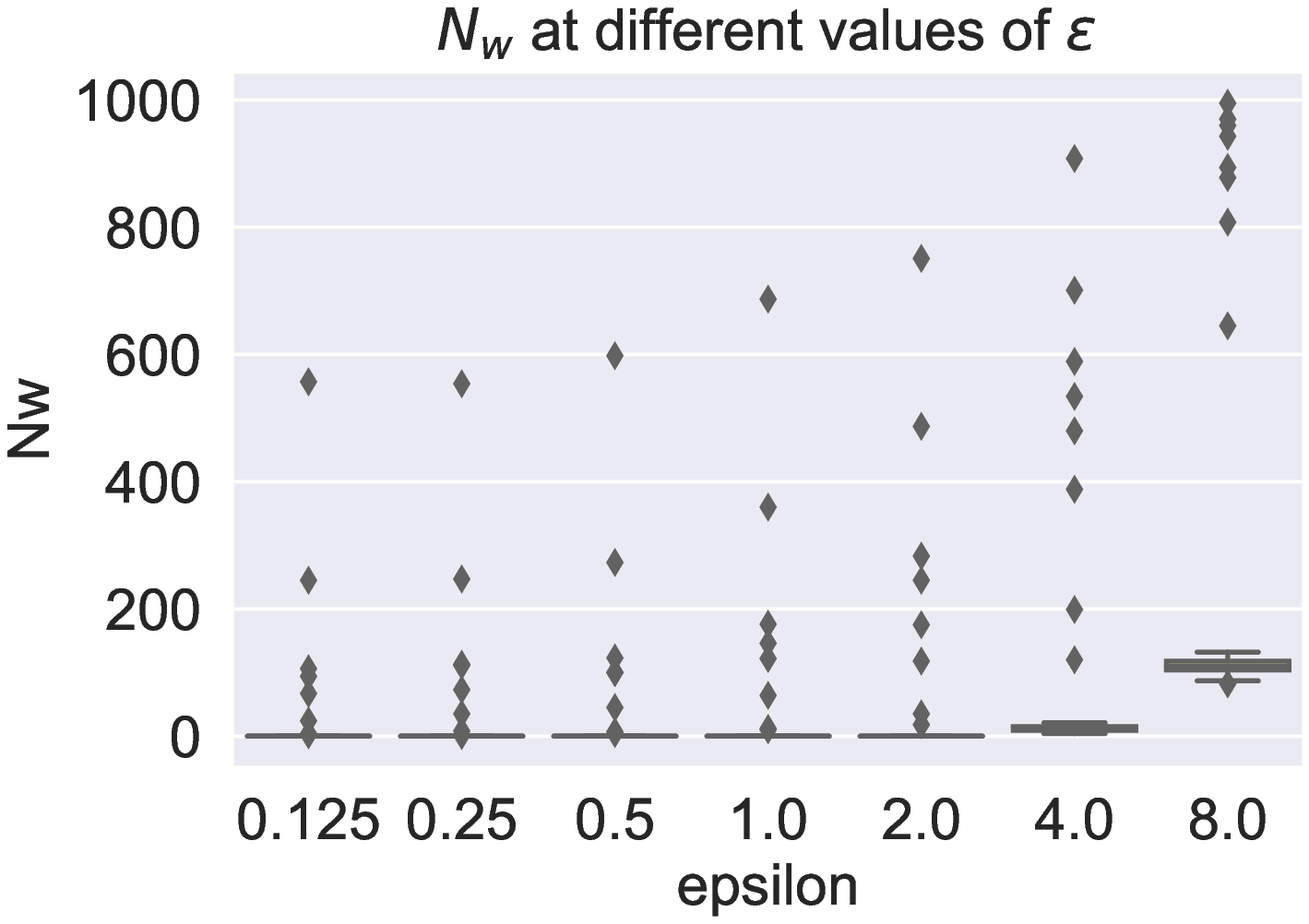}
    \caption{$N_w$ worst case}
    \label{fig:nw_stats_worst}
\end{subfigure}
  \begin{subfigure}[b]{0.24\textwidth}
    \includegraphics[width=\textwidth]{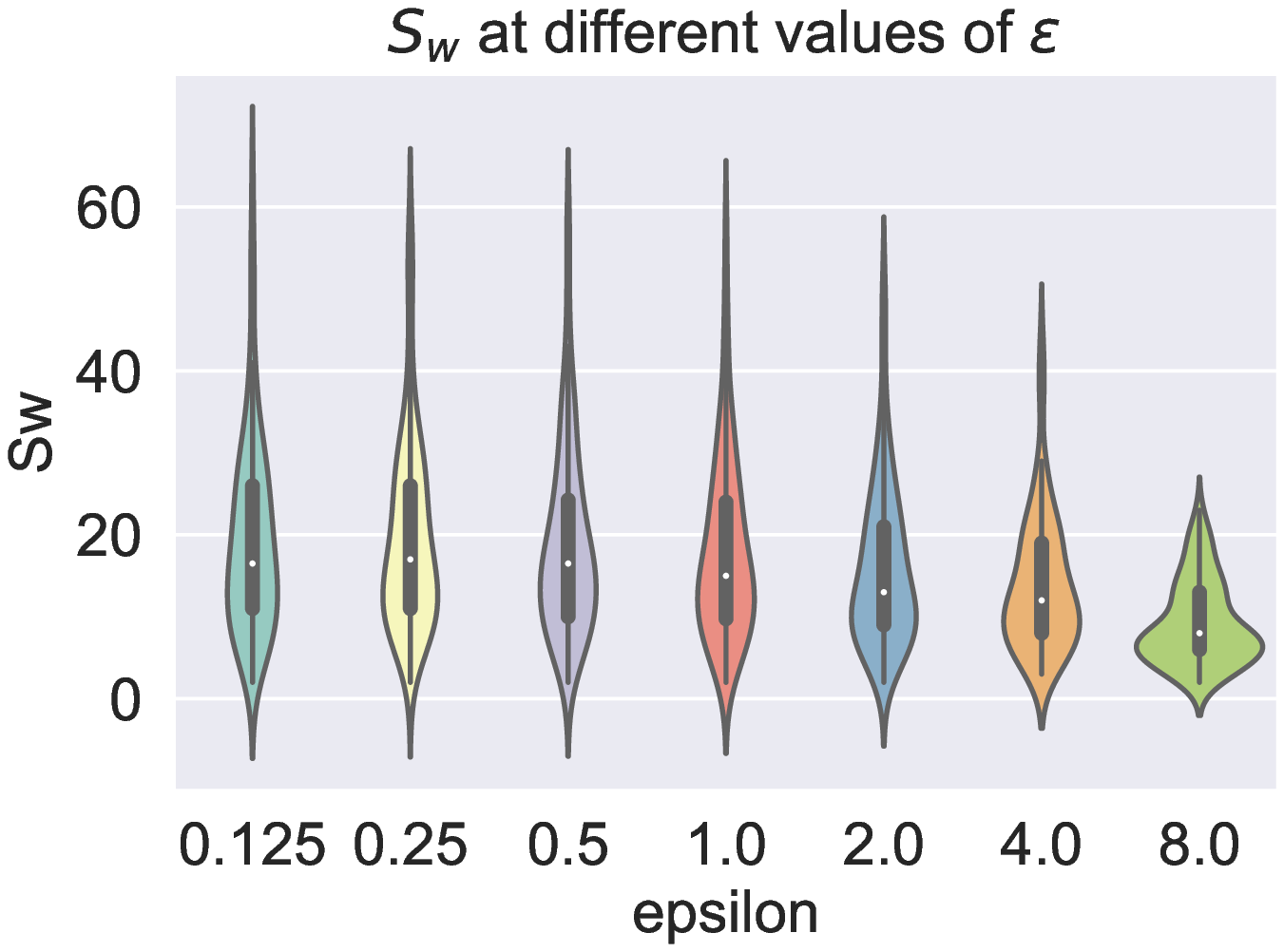}
    \caption{$S_w$ uncertainty statistics}
    \label{fig:sw_stats}
\end{subfigure}
  \begin{subfigure}[b]{0.25\textwidth}
    \includegraphics[width=\textwidth]{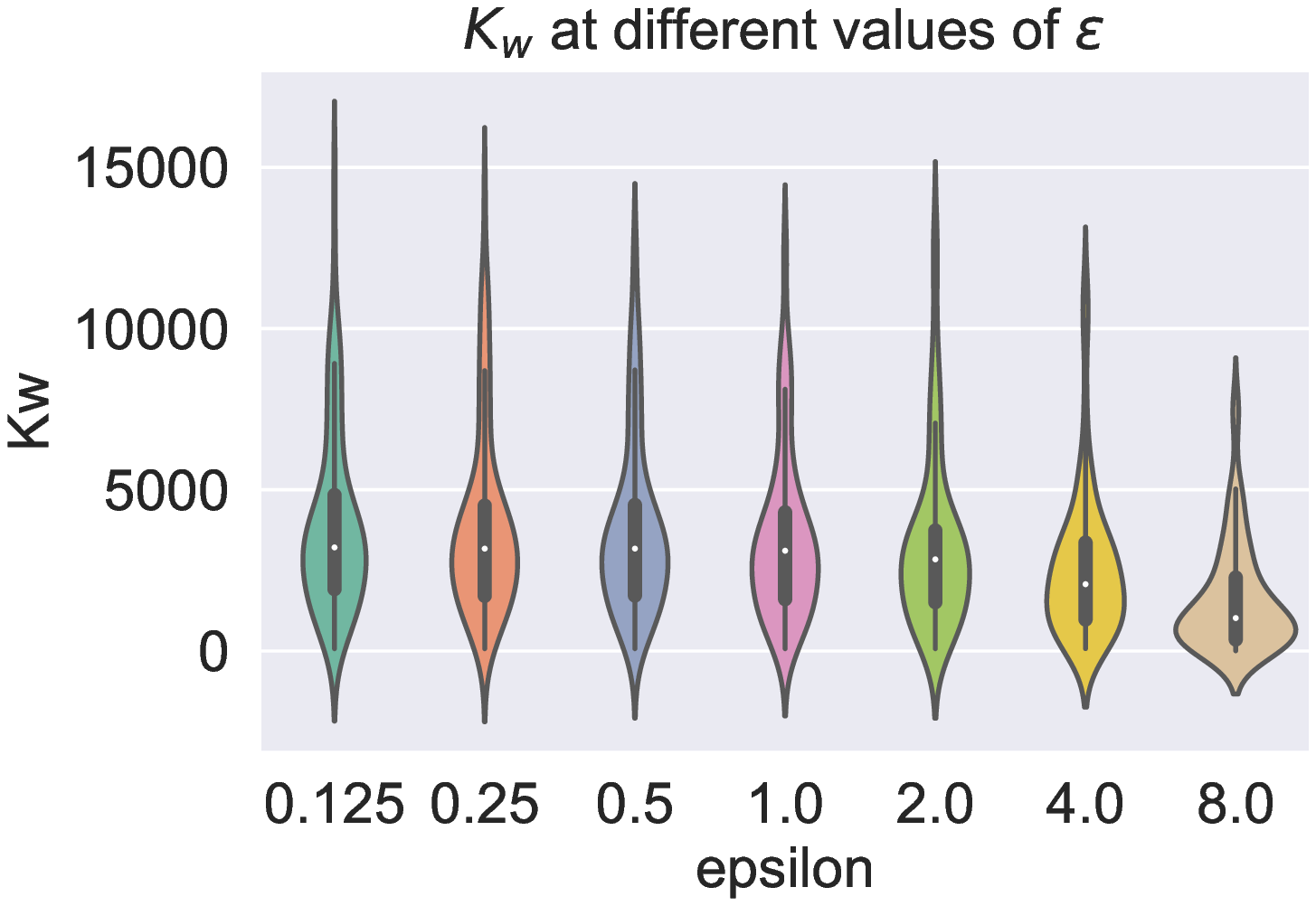}
    \caption{$K_w$ indistinguishability}
    \label{fig:kw_stats}
\end{subfigure}
    \caption{Privacy statistics -- (a) $N_w$ statistics: avg count of $M(w) = w$ (b) $N_w$ statistics: max count of $M(w) = w$ (c) $S_w$ statistics: distinct outputs for $M(w)$ (d) $K_w$ statistics: count of different words $|\{w, w'\}|$ which resolve to same output $\hat{w}$}
\label{fig:privacy_statistics}
\end{figure*}

We estimated the quantities $N_w$ and $S_w$ empirically by running the mechanism $M(w)$ $1,000$ times for a random population ($10,000$ words) of the vocabulary $\cW$ at different values of $\varepsilon$. The results are presented in Fig.~\ref{fig:nw_stats} and \ref{fig:nw_stats_worst} for $N_w$ and Fig.~\ref{fig:sw_stats} for $S_w$.

\subsection{Indistinguishability statistics}
Indistinguishability metrics of privacy indicate denote the ability of the adversary to distinguish between two items of interest. $d_{\chi}$-privacy provides degrees of indistinguishability of \emph{outputs} bounded by the privacy loss parameter $\varepsilon$. For example, given a query $x =$ `send the package to \emph{London}', corresponding outputs of $\hat{x} = $ `send the package to  \emph{England}' or `\ldots to \emph{Britain}' provide privacy by \emph{output} indistinguishability for the user. This is captured as uncertainty over the number of random new words as expressed in the $S_w$ metric. 

However, we also extend our privacy guarantees to indistinguishability over the \emph{inputs}. For example, for an adversary observing the output $\hat{x} = $ `send the package to \emph{England}', they are unable to infer the input $x$ that created the output $\hat{x}$ because, for the permuted word $\hat{w} = $ \emph{England}, the original word $w$ could have been any member of the set $\{w : \forall w \in \hat{\cW} \enspace \text{if} \enspace w \prec \hat{w} \}$, where $a \prec b$ implies that $a$ is lower than $b$ in the embedding hierarchy. For example, $\{ {\footnotesize \text{\sc{London}}, \text{\sc{Manchester}}, \text{\sc{Birmingham}}, \ldots }\} \prec \{ {\footnotesize \text{\sc{England}}, \text{\sc{Britain}}, \ldots }\}$. Since this new statistic derives from $S_w$, we expect it to vary across $\varepsilon$ in the same manner. Hence, we replace $\hat{\cW}$ with $S_w$ and define the new statistic $K_w$ as:
{\small
\begin{align*}
K_w = \min |\{w : \forall w \in S_w \enspace \text{if} \enspace w \prec \hat{w} \}|
\end{align*}
}
This input indistinguishability statistic can be thought of formally in terms of \emph{plausible deniability} \cite{bindschaedler2017plausible}. In \cite{bindschaedler2017plausible}, plausible deniability states that an adversary cannot deduce that a particular \emph{input} was significantly more responsible for an observed \emph{output}. This means, there exists a set of inputs that could have generated the given output with about the same probability. Therefore, given a vocabulary size $|\cW| > k$ and mechanism $M$ such that $\hat{w} = M(w_1)$, we get $k-$indistinguishability over the inputs with probability $\gamma$ if there are at least $k - 1$ distinct words $w_2, ..., w_k \in \cW \enspace$ \textbackslash $\{d_1\}$ such that:

{\small
\begin{align*}
\gamma^{-1} \leq \frac{\Pr[M(w_i)] = \hat{w}}{\Pr[M(w_j)] = \hat{w}} \leq \gamma \enspace \text{for any} \enspace i, j \in \{1, 2, ..., k\}
\end{align*}
}
We also estimated the values of $K_w$ empirically by running the mechanism $1,000$ times for a random population ($10,000$ words) of the vocabulary $\cW$ at different values of $\varepsilon$. The results are presented in Fig.~\ref{fig:kw_stats}.

\subsection{Selecting a value of $\varepsilon$}
To set the value of $\varepsilon$ for a given task, we propose following the guidelines offered by \cite{chatzikokolakis2015constructing} in the context of location privacy by providing appropriate reformulations. They suggest mapping $\varepsilon$ to a desired radius of \emph{high protection} within which, all points have the same distinguishability level. We can achieve a corresponding calibration using results in Fig.~\ref{fig:privacy_statistics}.

The worst-case guarantees highlighted by the upper bound of the $N_w$ statistic (see Fig.~\ref{fig:nw_stats_worst}) equips us with a way to fix an equivalent `radius of \emph{high protection}'. This `radius' corresponds to the upper bound on the probability $\Pr[M(w) = w]$ which sets the guarantee on the likelihood of changing \emph{any} word in the embedding vocabulary. Consequently, the words which are provided with the `same distinguishability level' can be interpreted by the size of the results in Fig.~\ref{fig:sw_stats}, and by extension, Fig.~\ref{fig:kw_stats}. In the following sections, we investigate the impact of setting varying values of $\varepsilon$ on the performance of downstream \ac{ML} models, and how the privacy guarantees of our Hyperbolic model compares to the Euclidean baseline.

\section{Experiments}
We carry out $3$ experiments to illustrate the tradeoff between privacy and utility. The first two are \emph{privacy} experiments, while the third is a set of \emph{utility} experiments against \ac{ML} tasks. 

\subsection{Evaluation metrics}
\begin{itemize}
\item \emph{Author predictions}: the number of authors that were re-identified in a dataset. Lower is better for privacy. 
\item \textbf{$N_w$}: number of times (of $1,000$) where the mechanism returned the original word. Lower is better for privacy.
\item \emph{Accuracy}: is the percentage of predictions the downstream model got right. Higher is better for utility
\end{itemize}

\subsection{Privacy experiments I}
In this section, we describe how we carried out privacy evaluations using an approach similar to \cite{fernandes2018generalised}.

\subsubsection{\textbf{Task}} 
The task is to carry out author \emph{obfuscation} against an authorship \emph{attribution} algorithm.

\subsubsection{\textbf{Baselines}} 
Just as in \cite{fernandes2018generalised}, the `adversarial' attribution algorithm is Koppel's algorithm \cite{koppel2011authorship}. Evaluation datasets were selected from the PAN@Clef tasks as follows:

\begin{itemize}
\item \textbf{PAN11} \cite{argamon2011overview} the \emph{small} dataset contained $3,001$ documents by $26$ authors while the \emph{large} set had $9,337$ documents by $72$ authors. Both were derived from the Enron emails.
\item \textbf{PAN12} \cite{juola2012overview} unlike the short email lengths per author in PAN11, this dataset consisted of dense volumes per author. Set-A had $3$ authors with between $1,800$ and $6,060$ words; C and D had $8$ authors with $\approx 13,000$ words each; while set-I consisted of $14$ authors of novels with word counts ranging between $40,000$ to $170,000$. 
\end{itemize}

\subsubsection{\textbf{Experiment setup}} 
We ran each dataset against Koppel's algorithm \cite{koppel2011authorship} to get the baseline. Each dataset was then passed through our $d_{\chi}$-privacy algorithm to get a new text output. This was done line by line in a manner similar to \cite{fernandes2018generalised} \ie{} all non stop words were considered. We evaluated our approach at the following values of $\varepsilon = 0.5, 1, 2$ and $8$.

\subsubsection{\textbf{Privacy experiment results}}
The results in Table~\ref{tab:priv_res_1} show that our algorithm provides tunable privacy guarantees against the authorship model. It also extends guarantees to authors with thousands of words. As $\varepsilon$ increases, the privacy guarantees decrease as clearly evidenced by the PAN11 tasks. We only show results for Kopell's algorithm because other evaluations perform worse on the baselines. \eg{} PAN18 identifies only $65$ and $50$ authors, while PAN19-SVM identifies $54$ and $35$ authors in the PAN11 small and large datasets.


\begin{table}[h]
\smaller
\begin{center}
\begin{tabular}{| l | R | R | R | R | R | R | l | }
 \hline
 & \multicolumn{2}{c|}{\textsc{Pan}-$11$} & \multicolumn{4}{c|}{\textsc{Pan}-$12$}\\
 \hline
$\varepsilon$ & \text{small} & \text{large} & \text{set-A} & \text{set-C} & \text{set-D} &  \text{set-I} \\ 
 \hline
 \hline
 $\mathbf{0.5}$ & 36 & 72 & 4 & 3 & 2 & 5\\
 $\mathbf{1}$ & 35 & 73 & 3 & 3 & 2 & 5\\
 $\mathbf{2}$ & 40 & 78 & 4 & 3 & 2 & 5\\
 $\mathbf{8}$ & 65 & 116 & 4 & 5 & 4 & 5\\
\hline
$\infty$ & 147 & 259 & 6 & 6 & 6 & 12\\
\hline
\end{tabular}
\caption{Correct author predictions (lower is better)} \label{tab:priv_res_1}
\end{center}
\end{table}

\subsection{Privacy experiments II}
We now describe how we evaluate the privacy guarantees of our Hyperbolic model against the Euclidean baseline.

\subsubsection{\textbf{Task and baselines}} 
The objective was to compare the expected privacy guarantees for our Hyperbolic \vs{} the Euclidean baseline, given the same worst case guarantees. We evaluated against $100d$, $200d$ and $300d$ GloVe embeddings.

\subsubsection{\textbf{Experiment setup}} 
We designed the $d_{\chi}$-privacy algorithm in the Euclidean space as follows: (a) the embedding model was GloVe, using the same vocabulary as in the \pc{} embeddings described in Sec.~\ref{sec:pc_embeddings}; (b) we sampled using the multivariate Laplacian distribution, by extending the planar Laplacian in \cite{andres2013geo} using the technique in \cite{wsdm_paper}; (c) we calibrate Euclidean $\varepsilon$ values by computing the privacy statistics $N_w$ for a given Hyperbolic $\varepsilon$ value.

To run the experiment, we repeat the following for the Hyperbolic and Euclidean embeddings: (1) first, we select a value of $\varepsilon$, (2) we empirically compute the worst case guarantee, \ie{} the largest maximum number of times we get \emph{any} word $\underset{w \in \cW}{\max}[M(w) = w]$ rather than selecting a new word after our noise perturbation, (3) we compute the expected guarantee, \ie{} the average number of times we get \emph{all} words each time we perturb the word $\underset{w \in \cW}{\text{avg}}[M(w) = w]$.

\subsubsection{\textbf{Privacy experiment results}}
The results for the comparative privacy analysis are presented in Tab.~\ref{tab:privacy_results}. The results clearly demonstrate that for identical worst case guarantees, the expected case for the Hyperbolic model is significantly better than the Euclidean across all Euclidean dimensions. Combining this with the superior ability of the Hyperbolic model to encode both similarity and hierarchy even at lower dimensions provides a strong argument for adopting it as a $d_{\chi}$-privacy preserving mechanism for the motivating examples described in Sec.~\ref{sec:mech_overview}.

\begin{table}[h]
\smaller
\begin{center}
\begin{tabular}{| l | c | r | r | r | r | }
 \hline
 & & \multicolumn{4}{c|}{expected value $N_w$} \\
 \hline
$\varepsilon$ & worst-case $N_w$ & \textsc{hyp}-$100$ & \textsc{euc}-$100$ & \textsc{euc}-$200$ & \textsc{euc}-$300$ \\ 
 \hline
 \hline
 $\mathbf{0.125}$ & $134$ & \textbf{1.25} & $38.54$ & $39.66$ & $39.88$ \\
 $\mathbf{0.5}$ & $148$ & \textbf{1.62} & $42.48$ & $43.62$ & $43.44$ \\
 $\mathbf{1}$ & $172$ & \textbf{2.07} & $48.80$ & $50.26$ & $53.82$ \\
 $\mathbf{2}$ & $297$ & \textbf{3.92} & $92.42$ & $93.75$ & $90.90$ \\
 $\mathbf{8}$ & $960$ & \textbf{140.67} & $602.21$ & $613.11$ & $587.68$ \\
  
\hline
\end{tabular}
\caption{Privacy comparisons (lower $N_w$ is better)} \label{tab:privacy_results}
\end{center}
\end{table}

\subsection{Utility experiments}
Having established Hyperbolic embeddings as being better than the Euclidean baseline for $d_{\chi}$-privacy, we now demonstrate its effects on the utility of downstream models (\ie{} we conduct utility experiments \emph{only} on Hyperbolic embeddings).

\subsubsection{\textbf{\ac{ML} Tasks}}

we ran experiments on $8$ tasks ($5$ classification and $3$ natural language tasks) to highlight the tradeoff between privacy and utility for a broad range of tasks. See Tab.~\ref{tab:classification_tasks} for a summary of the tasks and datasets. \\

\begin{table}[h]
\smaller
\begin{center}
\begin{tabular}{ | l | R | l | L | l | }
 \hline
 \textbf{name} & \textbf{samples} & \textbf{task} & \textbf{classes} & \textbf{example(s)} \\ 
 \hline
MR \cite{pang2005seeing} & 10,662 & sentiment (movies) & 2 & neg, pos \\
CR \cite{hu2004mining} & 3,775 & product reviews & 2 & neg, pos \\
MPQA \cite{wiebe2005annotating} & 10,606 & opinion polarity & 2 & neg, pos \\
SST-5 \cite{socher2013recursive} & 12,000 & sentiment (movies) & 5 & 0 \\
TREC-6 \cite{li2002learning} & 5,452 & question-type & 6 & LOC:city \\
\hline
SICK-E \cite{marelli2004sick} & 10,000 & natural language inference & 3 & contradiction \\
MRPC \cite{dolan2004unsupervised} & 5,801 & paraphrase detection & 2 & paraphrased \\
STS14 \cite{agirre2014semeval} & 4,500 & semantic textual similarity & [0,5] & 4.6\\
\hline
\end{tabular}
\caption{Classification and natural language tasks} \label{tab:classification_tasks}
\end{center}
\end{table}

%
%
%
%
%
%
%
%
%
%

\subsubsection{\textbf{Task baselines}} 
the utility results were baselined using {SentEval} \cite{conneau2018senteval}, an evaluation toolkit for sentence embeddings. We evaluated the utility of our algorithm against an upper and lower bound.

To set an upper bound on utility, we ran each task using the original datasets. Each task was done on the following embedding representations: (1) {InferSent} \cite{conneau2017supervised}, (2) {SkipThought} \cite{kiros2015skip} and (3) {fastText} \cite{bojanowski2016enriching} (as an average of word vectors). 

To set a lower bound on the utility scores, rather than replacing words using our algorithm, we replaced them with \emph{random} words from the embedding vocabulary.

\subsubsection{\textbf{Experiment setup}} unlike intent classification datasets such as \cite{coucke2018snips} and \cite{tur2010left}, most datasets do not come with `slot values' to be processed by a privacy preserving algorithm (see motivating examples in Sec.~\ref{sec:mech_overview}). As a result, we pre-processed all the datasets to identify phrases with \emph{low transition probabilities} using the privacy preserving algorithm proposed by \cite{chen2012differentially}.

The output from \cite{chen2012differentially} yields a sequence of high frequency sub-phrases. As a result, for every query in each dataset, we are able to (inversely) select a set of low transition phrases which act as slot values to be fed into our algorithm.

For a given dataset, the output from processing each query using our algorithm is then fed into the corresponding task.

\subsubsection{\textbf{Utility experiment results}}
We evaluated our algorithm at values of $\varepsilon = 0.125, 1$ and $8$. Words were sampled from the metric space defined by the $100d$ \pc{} embeddings described in Sec~\ref{sec:pc_embeddings}.

The results are presented in Tab.~\ref{tab:all_utility_results}. The evaluation metric for all tasks was accuracy on the test set. 
Across all the experiments, our algorithm yielded better results that just replacing with random words.
In addition, and as expected, at lower values of $\varepsilon = 0.125$, we record lower values of utility across all tasks. Conversely at the higher value of $\varepsilon = 8$, the accuracy scores get closer to the baselines.
All the results illustrate the tradeoff between privacy and utility. It also shows that we can achieve tunable privacy guarantees with minimal impact on the utility of downstream \ac{ML} models.

\begin{table*}[h]
\smaller
\begin{center}
\begin{tabular}{| l | L | l | l | l | L | L | L | }
 \hline
 & & \multicolumn{3}{c|}{\textsc{hyp}-$100d$} & \multicolumn{3}{c|}{\emph{original}}\\
 \hline
 \textbf{dataset} & \emph{random} & $\mathbf{\varepsilon = 0.125}$ & $\mathbf{\varepsilon = 1}$ & $\mathbf{\varepsilon = 8}$ & \text{InferSent} & \text{SkipThought} & \text{fastText-BoV}  \\ 
 \hline
 \hline
MR & 58.19 & $58.38$ & $63.56$ & $74.52$ & 81.10 & 79.40 & 78.20 \\
CR & 77.48 & $83.21^{**}$ & $83.92^{**}$ & $85.19^{**}$ & 86.30 & 83.1 & 80.20 \\
MPQA & 84.27 & $88.53^*$ & $88.62^*$ & $88.98^*$ & 90.20 & 89.30 & 88.00 \\
SST-5 & 30.81 & $41.76$ & $42.40$ & $42.53$ & 46.30 & - & 45.10 \\
TREC-6 & 75.20 & $82.40$ & $82.40$ & $84.20^*$ & 88.20 & 88.40 & 83.40 \\
\hline
SICK-E & 79.20 & $81.00^{**}$ & $82.38^{**}$ & $82.34^{**}$ & 86.10 & 79.5 & 78.9 \\
MRPC & 69.86 & $74.78^*$ & $75.07^*$ & $75.01^*$ & 76.20 & - & 74.40 \\
STS14 & 0.17/0.16 & $0.44/0.45$ & $0.45/0.46^*$ & $0.52/0.53^*$ & 0.68/0.65 & 0.44/0.45 & 0.65/0.63 \\
\hline
\end{tabular}
\caption{Accuracy scores on classification tasks. * indicates results better than $1$ baseline, ** better than $2$ baselines} \label{tab:all_utility_results}
\end{center}
\end{table*}



\section{Related work}
There are two sets of research most similar to ours. The recent work by \cite{wsdm_paper} and \cite{fernandes2018generalised} applies $d_{\chi}$-privacy to text using similar techniques to ours. However, their approach was done in Euclidean space while ours used word representations in, and noise sampled from Hyperbolic space. As a result, our approach can better preserve semantics at smaller values of $\varepsilon$ by selecting hierarchical replacements.

The next set include research by \cite{cumby2011machine}, \cite{sanchez2016c}, and \cite{anandan2012t}. These all work by identifying sensitive terms in a document and replacing them by some generalization of the word. This is similar to what happens when we sample from Hyperbolic space towards the center of the \pc{} ball to select a hypernym of the current word. The difference between these and our work is the mechanism for selecting the words and the privacy model used to describe the guarantees provided.

\section{Conclusion}
This paper is the first to demonstrate how hierarchical word representations in Hyperbolic space can be deployed to satisfy $d_{\chi}$-privacy in the text domain. We presented a theoretical proof of the privacy guarantees in addition to defining a probability distribution for sampling privacy preserving noise from Hyperbolic space. Our experiments illustrate that our approach preserves privacy against an author attribution model and utility on several downstream models. Compared to the Euclidean baseline, we observe $> 20$x greater guarantees on expected privacy against comparable worst case statistics. Our results significantly advance the study of $d_{\chi}$-privacy, making generalized differential privacy with provable guarantees closer to practical deployment in the text domain.

\bibliographystyle{IEEEtran}
\bibliography{IEEEabrv,madlib_icdm}

\end{document}